\documentclass[10pt,twocolumn,letterpaper]{article}

\usepackage{cvpr}
\usepackage{times}
\usepackage{epsfig}
\usepackage{graphicx}
\usepackage{amsmath}
\usepackage{amssymb}
\usepackage{multirow}
\usepackage{float}
\usepackage{caption} 
\usepackage{textcomp}
\makeatletter
\newcommand{\printfnsymbol}[1]{%
  \textsuperscript{\@fnsymbol{#1}}%
}
\makeatother

\usepackage[pagebackref=true,breaklinks=true,letterpaper=true,colorlinks,bookmarks=false]{hyperref}

\cvprfinalcopy % *** Uncomment this line for the final submission

\def\cvprPaperID{102} % *** Enter the CVPR Paper ID here

\ifcvprfinal\fi
\begin{document}

%%%%%%%%% TITLE
\title{Fast Deep Multi-patch Hierarchical Network for Nonhomogeneous Image Dehazing}

\author{Sourya Dipta Das\thanks{Equal contribution.}\\
Jadavpur University\\
Kolkata, India\\
{\tt\small dipta.juetce@gmail.com}
\and
Saikat Dutta\printfnsymbol{1} \\
 IIT Madras\\
 Chennai, India\\
{\tt\small cs18s016@smail.iitm.ac.in}
}

\maketitle
%\thispagestyle{empty}

%%%%%%%%% ABSTRACT
% Image Dehazing - challenging problem
% Short importance of non-homog. dehazing
% We introduce a FAST network for dehazing suitable for real-time 
% suitable for dense haze removal too
\begin{abstract}
   Recently, CNN based end-to-end deep learning methods achieve superiority in Image Dehazing but they tend to fail drastically in Non-homogeneous dehazing. Apart from that, existing popular Multi-scale approaches are runtime intensive and memory inefficient. In this context, we proposed a fast Deep Multi-patch Hierarchical Network to restore Non-homogeneous hazed images by aggregating features from multiple image patches from different spatial sections of the hazed image with fewer number of network parameters. Our proposed method is quite robust for different environments with various density of the haze or fog in the scene and very lightweight as the total size of the model is around 21.7 MB. It also provides faster runtime compared to current multi-scale methods with an average runtime of 0.0145s to process $1200 \times 1600$ HD quality image. Finally, we show the superiority of this network on Dense Haze Removal to other state-of-the-art models.
\end{abstract}

%%%%%%%%% BODY TEXT
\section{Introduction}
Outdoor images are often deteriorated due to the extreme weather, such as fog and haze, which influences visibility issues in the scene because of the degradation of color, contrast and textures for different distant objects, selective attenuation of the light spectrum. Restoring such hazed images has become an important problem in many computer vision applications like visual surveillance, remote sensing, and Autonomous transportation etc. Most of early methods proposed for image dehazing are based on the classic atmospheric scattering model which is shown as the following equation. \ref{scattering}.
\begin{equation}
\label{scattering}
    I(x) = J(x)t(x) + A(1 - t(x))
\end{equation}
where, $x$ represents pixel locations, $I(x)$ is the observed hazy image, $J(x)$ is the dehazed image, $t(x)$ is called medium transmission function and $A$ is the global atmospheric light.
Recently, Deep learning based methods have shown remarkable improvements though those methods suffer from degradation of colour, texture in image, halo artifacts, haze residuals and distortions. In our problem statement, Non-homogeneous haze in the scene can be seen in the real world situation where different spatial domains of the image can be affected by different levels of haze. The degradation level also vary a lot for objects at different scene depth due to non-uniform haze distribution in the image.  
Few example images of such Non-homogeneous haze are shown in figure \ref{nh_haze}. Dehazing model should put more effort to handle non-uniform haze and different degradation between different scene depth jointly. Multi-scale and scale-recurrent models can be a viable solution in this type of problem because of its coarse-to-fine learning scheme by hierarchical integration of features from different spatial scale of the image. This type of methods is inefficient because of high runtime and large model size due to a lot of convolution and Deconvolution layers. Apart from that, increasing depth of layers at fine scale levels may not always improve the perceptual quality of the output dehazed image. On the contrary, main goal of our model is to aggregate features multiple image patches from different spatial sections of the image for better performance. The parameters of our encoder and decoder are very less due to residual links in our model which helps in fast dehazing inference. The main intuition behind our idea is to make the lower level network portion focus on local information by extracting local features from the finer grid to produce residual information for the upper level part of the network to get more global information from both finer and coarser grid which is achieved by concatenating convolutional features. 

%-------------------------------------------------------------------------
\section{Related Work}
Most early work of image dehazing methods is developed on atmosphere scattering model as it's physical model. In that respect, previous works on image dehazing can be segregated into two classes which are traditional image prior-based methods and end to end deep learning based methods. Traditional image prior based methods relies on hand-crafted statistics from the images to leverage extra mathematical constraints to compensate for the information lost during reconstruction. On contrary, deep learning based methods learn the direct relationship between haze and haze-free image by utilizing multistage, attention mechanisms etc. Here, we discussed some recent deep learning based methods with state-of-the-art results. 

Zhang et al.\cite{zhangdensely} proposed a dehazing network with edge-preserving
densely connected encoder-decoder architecture that jointly learns the dehazed image, transmission map and atmosphere light all together based on the scattering model for dehazing. In their encoder-decoder architecture, they use a multilevel pyramid pooling module and to improve their results further, joint-discriminator based on GAN is used to incorporate the correlation between estimated transmission map and dehazed image. Deng et al.\cite{deng2019deep} presents a multi-model fusion network to combine multiple models in its different levels of layers and enhance the overall performance of image dehazing. They generate the multi-model attention integrated feature from various CNN features at different levels and fed it to their fusion model to predict dehazed image for an atmospheric scattering model and four haze-layer separation models altogether. After that, they fused the corresponding results together to generate the final dehazed image. Qin et al.\cite{qin2019ffa} proposed a novel Feature Attention module which fuses Channel Attention with Pixel Attention while considering different
weighted information of different channel-wise features and uneven haze distribution on different pixels of the hazed image. For Outdoor hazy images, their work proves superiority though it didn't work well in case of dense dehazing. Liu et al.\cite{liu2019griddehazenet} proposed a grid network with attention-based multi-scale estimation which overcomes the bottleneck problems found in general multi-scale approach. Apart from that, their method also consists of pre-processing and post-processing modules. The pre-processing module used in this method is trainable to get more relevant features from diversified pre-processed image inputs and it outperforms the other hand picked classical pre-process techniques. The post-processing module is finally used on intermediate dehazed image to get more finer dehazed image. Their study shows how their method works quite independently and does not take any advantage from atmosphere scattering model for image dehazing.

Unlike other multi-stage methods, Li et al.\cite{li2019lap} used a level aware progressive deep network to learn different levels of haze from its different stages of the network by different supervision. Their network tends to progressively learn gradually more intense haze from image by focusing on a specific part of image with a certain haze level. They have also devised a adaptive hierarchical integration technique by cooperating with the it's memory network component and domain information of dehazing to emphasize the well-reconstructed parts of the image in it's each stage of the network. Liu et al.\cite{liu2019learning} suggests a method to learn a haze relevant image priors by using a iteration algorithm with deep CNNs. They achieve this by using gradient descent method to optimize a variational model with image fidelity terms and proper regularization. this method indeed a great combination of properties from classical deep learning based method and physical hazed image formation model. Sharma et al.\cite{sharma2020scale} explored the application of Laplacians of Gaussian
(LoG) of the images to reattain the edge and intensity variation information. They optimize their end-to-end  deep model by per-pixel difference between Laplacians of Gaussians of the dehazed and ground truth images. they additionally do adversarial training with a perceptual loss to enhance their results. Apart from other physical scattering model based methods, GAN , multiscale or multistage deep networks, Image dehazing can also be posed as image to image translation problem. Qu et al.(2019)\cite{qu2019enhanced} proposed their solution as an enhanced Pix2Pix Model which is widely used in image style transfer, image to image translation etc. problems. Their method consists of a GAN with a Enhancer modules to support the dehazing process to get more detailed, vivid image with less artifacts. Their work also proved superiority over other methods in the aspect of the perceptual quality of the dehazed images.

\section{Proposed Method}
We use a Multi-patch and a Multi-scale network for Non-homogeneous Image Dehazing. In this section, we describe these two architectures in detail.
\subsection{Multi-patch Architecture:} We use Deep Multi-patch Hierarchical Network(DMPHN). DMPHN is originally used for Single Image Deblurring\cite{dmphn}. We use (1-2-4) variant of DMPHN in this paper. For the sake of completeness, we will discuss the architecture in the following.
\begin{figure*}[!h]
    \centering
    \includegraphics[width=0.7\textwidth]{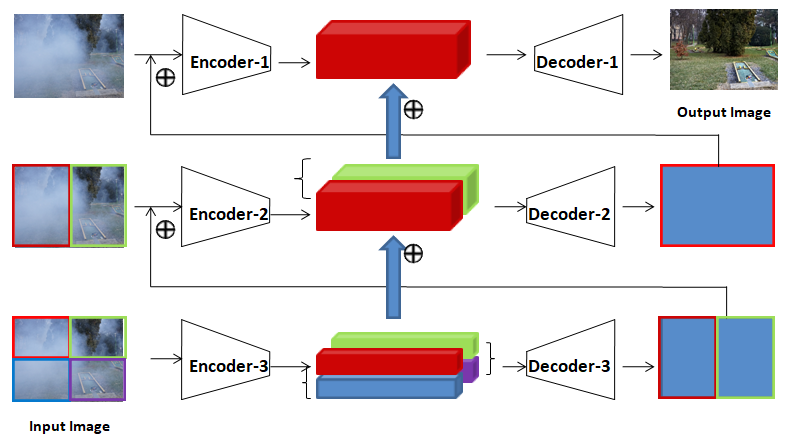}
    \caption{Architecture diagram of Deep Multi-Patch Hierarchical Network. `$\{$' denotes spatial concatenation and $\bigoplus$ denotes residual addition.}
    \label{mp_arch}
\end{figure*}

DMPHN is a multi-level architecture. There is an encoder-decoder pair in each level. Each level works on different number of patches. In DMPHN(1-2-4), the number of patches used is 1,2 and 4 from top to bottom levels respectively. The top-most level (level-1) considers only one patch per image. In the next level(level-2), the image is divided into two patches vertically. In the bottom-most level(level-3) the patches from previous level are further divided horizontally, resulting in total 4 patches.  

Let us consider an input hazy image $I^H$. We denote $j$-th patch in $i$-th level as $I^H_{i,j}$. In level-1, $I^H$ is not divided into any patches. In level-2, $I^H$ is divided vertically into $I^H_{2,1}$ and $I^H_{2,2}$. In level-3, $I^H_{2,1}$ and $I^H_{2,2}$ are divided horizontally to create 4 patches, $I^H_{3,1}, I^H_{3,2},I^H_{3,3}$ and $I^H_{3,4}$. Encoders and Decoders at $i$-th level is denoted as $Enc_i$
and $Dec_i$ respectively.

The information flow in DMPHN is bottom-up. Patches in the lowest level are fed to encoder $Enc_3$ to generate corresponding feature maps.
\begin{equation}
    F_{3,j} = Enc_i(I^H_{3,j}), \forall j \in [1,4]
\end{equation}
We concatenate spatially adjacent feature maps to obtain a new feature representation.
\begin{equation}
    P_{3,j} = [F_{3,2j-1},F_{3,2j}], \forall j\in [1,2]
\end{equation}
where [.,.] stands for concatenation. 

The new concatenated features are passed through decoder $Dec_3$.
\begin{equation}
    Q_{3,j} = Dec_3(P_{3,j}), \forall j\in [1,2]
\end{equation}
The decoder output is added with patches in the next level and fed to encoder.
\begin{equation}
    F_{2,j} = Enc_2(I^H_{2,j}+Q_{3,j}), \forall j\in [1,2]
\end{equation}
The encoder outputs are added with respective decoder inputs from previous level. Then the resulting feature maps are spatially concatenated.
\begin{align}
    F^*_{2,j} = F_{2,j} + P_{3,j}, \forall j\in [1,2] \\
    P_{2} = [F^*_{2,1},F^*_{2,2}]
\end{align}
$P_{2}$ is then fed to $Dec_2$ to generate residual feature maps for level-2.
\begin{equation}
    Q_{2} = Dec_2(P_{2})
\end{equation}
Decoder output at level-2 is added to input image and passed through $Enc_1$. Encoder output $F_1$ is added with decoder output at level-2, $Q_2$.
\begin{equation}
    F_{1} = Enc_1(I^H + Q_2)
\end{equation}
$F_{1}$ is added with $P_2$ and fed to $Dec_1$ to produce the final dehazed output $\hat{I}$.
\begin{align}
    P_1 = F_1 + P_2 \\
    \hat{I} = Dec_1(P_1)
\end{align}

\begin{figure*}[!h!t!p]
    \centering
    \includegraphics[width=0.7\textwidth]{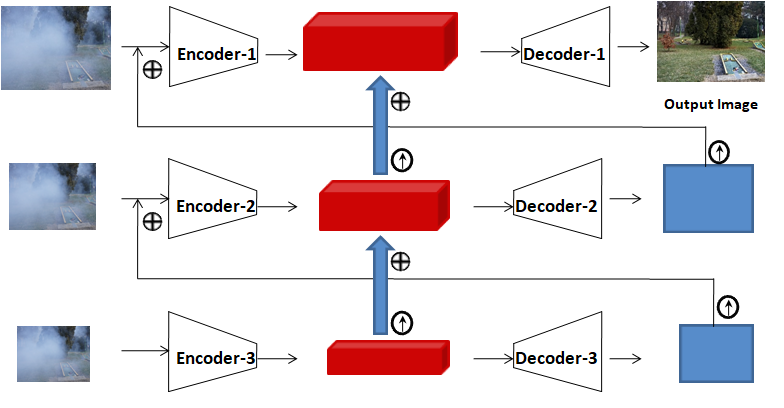}
    \caption{Architecture diagram of Deep Multi-Scale Hierarchical Network. \hspace{0.5\wd1}
 \setbox1=\hbox{\Huge $\circ$}%
\setbox2=\hbox{$i_3$}%
\makebox[0pt][c]{\usebox1}% Place \circlearrowright
\makebox[0pt][c]{\raisebox{2.5pt}{\footnotesize$\uparrow$}}% Overlay label
\hspace{0.5\wd1} denotes Upsampling by factor of 2 and $\bigoplus$ denotes residual addition.}
    \label{ms_arch}
\end{figure*}
\subsection{Multi-scale Architecture:} We also experiment with a multi-scale architecture. We name this architecture Deep Multi-scale Hierarchical Network(DMSHN). The details of the architecture are described as follows.

Input hazy image $I^H$ is downsampled by factor of 2 and 4 to create an image pyramid. We call these downsampled images $I^H_{0.5}$ and $I^H_{0.25}$ respectively. The architecture consists of 3 levels where each level has a pair of encoder and decoder. Encoder and decoder at level $i$ is denoted as $Enc_i$ and $Dec_i$ respectively.

At the lowest level $I^H_{0.25}$ is fed to encoder $Enc_3$ to obtain feature map $F_3$ and is further passed through decoder $Dec_3$ to feature representation $P_3$.
\begin{align}
    F_3 = Enc_3(I^H_{0.25}) \\
    P_3 = Dec_3(F_3)
\end{align}
$P_3$ is upscaled by factor of 2 and added to $I^H_{0.5}$ and passed through encoder $Enc_2$ to generate $F^*_2$. Encoder output from previous level is upscaled and added to intermediate feature map $F^*_2$ and fed to the decoder $Dec_2$.
\begin{align}
    F^*_2 = Enc_2(I^H_{0.5}+up(P_3)) \\
    F_2 = F^*_2 + up(F_3) \\
    P_2 = Dec_2(F_2)
\end{align}
where $up(.)$ denotes Upsampling operation by a factor of $2$.
Residual feature map $P_2$ from level-2 is added to the input hazy image and fed to encoder $Enc_1$. Encoder output is added with upscaled $F_2$ and passed through decoder to synthesize the dehazed output $\hat{I}$.
\begin{align}
    F^*_1 = Enc_2(I^H+up(P_2)) \\
    F_1 = F^*_1 + up(F_2) \\
    \hat{I} = Dec_1(F_1)
\end{align}
\subsection{Encoder and Decoder Architecture:}
We use the same encoder and decoder architecture at all levels of DMPHN and DMSHN. The
encoder consists of 15 convolutional layers, 6 residual connections and 6 ReLU units. The layers in the decoder and encoder are similar except that
2 convolutional layers are replaced by deconvolutional layers to generate
dehazed images as output.
\begin{figure*}[!h!t!p]
    \centering
    \includegraphics[width=\textwidth]{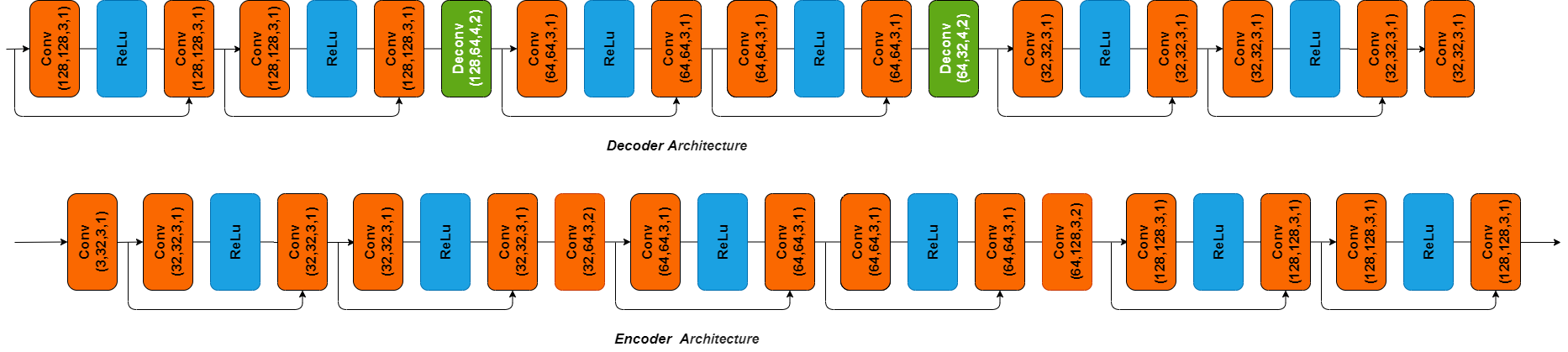}
    \caption{Encoder and Decoder architecture. Within brackets written values are Input Channel,
Output Channel,Kernel and Stride respectively.}
    \label{enc_dec_arch}
\end{figure*}
\section{Experiments}
 \subsection{Dataset Description: }
\begin{figure*}[!htbp]
    \centering
    \includegraphics[width=0.8\textwidth]{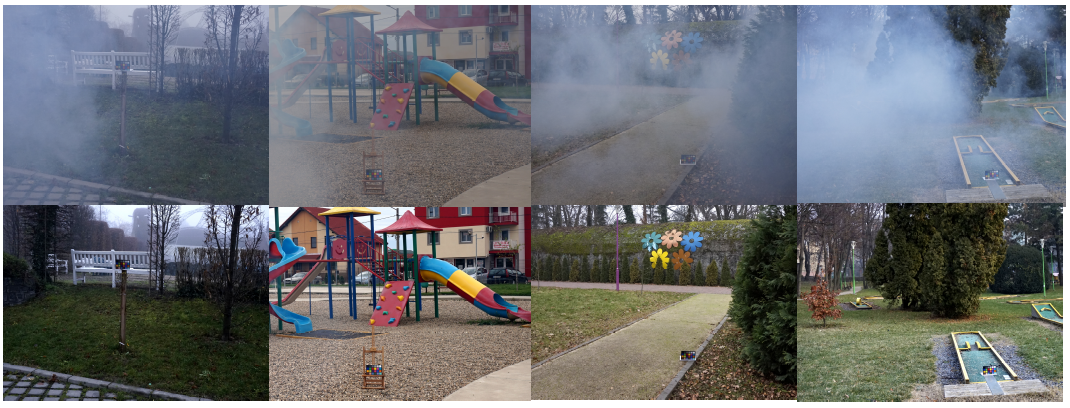}
    \caption{A snapshot of Training Dataset. Top row contains hazy images and bottom row contains corresponding ground truth images.}
    \label{nh_haze}
\end{figure*}

We used NH-HAZE dataset\cite{Ancuti_NH-HAZE_2020} provided for NTIRE 2020 Nonhomogeneous Image Dehazing challenge in our experiments. This dataset contains a total of 55 hazy and clear image pairs, divided into  trainset of 45 image pairs, validation set of 5 image pairs and test set of 5 image pairs. Validation and test ground truth images are not publicly available at this moment. Resolution of images in this dataset is $1200\times1600$. This dataset contains hazed and hazefree images of various outdoor scenes. A few hazefree and hazed image pairs from this dataset is shown in Figure-\ref{nh_haze}.
\subsection{Training data preparation:}
Due to the small amount of available data, we divide each image into 100 non-overlapping patches. Thus we obtain a training set of 4500 image-pairs of resolution $120\times160$. No data augmentation techniques were used.

\subsection{Loss functions:}
We use a linear combination of the following loss functions as our optimization objective.

\textbf{Reconstruction loss:} Reconstruction loss helps the network to generate dehazed frames close to the ground truth. Our reconstruction loss is a weighted sum of MAE or $L_1$ loss and MSE or $L_2$ loss. The reconstruction loss is given by,
\begin{equation}
    L_r = \lambda_{1}L_1 +  \lambda_{2}L_2
\end{equation}
where $L_1 = \left\|\hat{{I}}-I\right\|_{1}$ and $L_2 = \left\|\hat{{I}}-I\right\|_{2}$

\textbf{Perceptual loss:} $L_2$ distance between features extracted from conv4\_3 layer of VGGNet\cite{vggnet} of predicted and ground truth images are used as Perceptual loss\cite{johnson2016perceptual}. Perceptual loss is given by,
\begin{equation}
    L_p = \left\|\phi(\hat{{I}})-\phi(I)\right\|_{2}
\end{equation}

\textbf{TV loss:} We use Total Variation(TV) loss\cite{johnson2016perceptual} makes predictions smooth. TV loss is given by,
\begin{equation}
    L_{tv} = 
    \left\| \nabla_x \hat{I}\right\|_{2} + \left\| \nabla_y \hat{I}\right\|_{2}
\end{equation}

Our final loss function is given by, 
\begin{equation}
    L = \lambda_rL_r + \lambda_pL_p + \lambda_{tv}L_{tv} 
\end{equation}

In our experiments we choose $\lambda_r = 1$, $\lambda_p = 6e-3$, $\lambda_{tv} = 2e-8$. $\lambda_1$ and $\lambda_2$ is chosen to be $0.6$ and $0.4$ respectively.

\subsection{Training details:} 
We developed our models using Pytorch\cite{NEURIPS2019_9015} on a system with AMD Ryzen 1600X CPU and NVIDIA GTX 1080 GPU.
We use Adam optimizer\cite{kingma2014adam} to train our networks with values of $\beta_1$ and $\beta_2$ 0.9 and 0.99 respectively. We use batchsize of 8. Initial learning rate is set to be 1e-4 which is gradually decreased to 5e-5. We train our models until convergence.

\subsection{Testing details:}
We test our models' performance on the given full resolution images of validation data. Please note that, our models are fully convolutional, hence difference between train and test image size should not matter.

\subsection{Results:}
\subsubsection{Quantitative and Qualitative Results:}
As ground truth for validation set is not publicly available, we submit our validation results to Codalab server. We compare performance of our models with three state-of-the-art dehazing models namely AtJ-DH\cite{atjdh}, 123-CEDH\cite{123cedh} and FFA-Net\cite{qin2019ffa}. The quantitative results on Validation set are given in Table-\ref{quant_res}. DMPHN is performing better than the rest of the models. It can be observed that our Multi-patch network is performing better than our Multi-scale network in terms of both PSNR and SSIM. At lower levels of DMPHN, the network works on patch level, so the network learns local features compared to global features learnt by DMSHN, which explains the performance gain in DMPHN. 

Apart from decent dehazing results, it is to be noted that both DMPHN and DMSHN are lightweight and efficient models. Checkpoints of both the networks take 21.7 MB on disk. GPU processing times for DMPHN and DMPSN make them suitable for real-time applications.
\begin{table}[!h!t!p]
\centering
\begin{tabular}{|c|c|c|c|}
\hline
      & PSNR  & SSIM   & Runtime(s) \\ \hline
      AtJ-DH\cite{atjdh} & 15.94 & 0.5662 &   0.0775      \\ \hline
      123-CEDH\cite{123cedh} & 14.59 & 0.5488 &   0.0559      \\ \hline
      FFA-Net\cite{qin2019ffa} & 10.43 & 0.4168 &   1.7472      \\ \hline
DMPHN & \textbf{16.94} & \textbf{0.6177} &   \textbf{0.0145}      \\ \hline
DMPSN & 16.42 & 0.5991 &    0.0210     \\ \hline
\end{tabular}
\caption{Quantitative results on NH-HAZE\cite{Ancuti_NH-HAZE_2020} Validation set.}
\label{quant_res}
\end{table}
\begin{figure*}[!h]
    \centering
    \includegraphics[width=\textwidth]{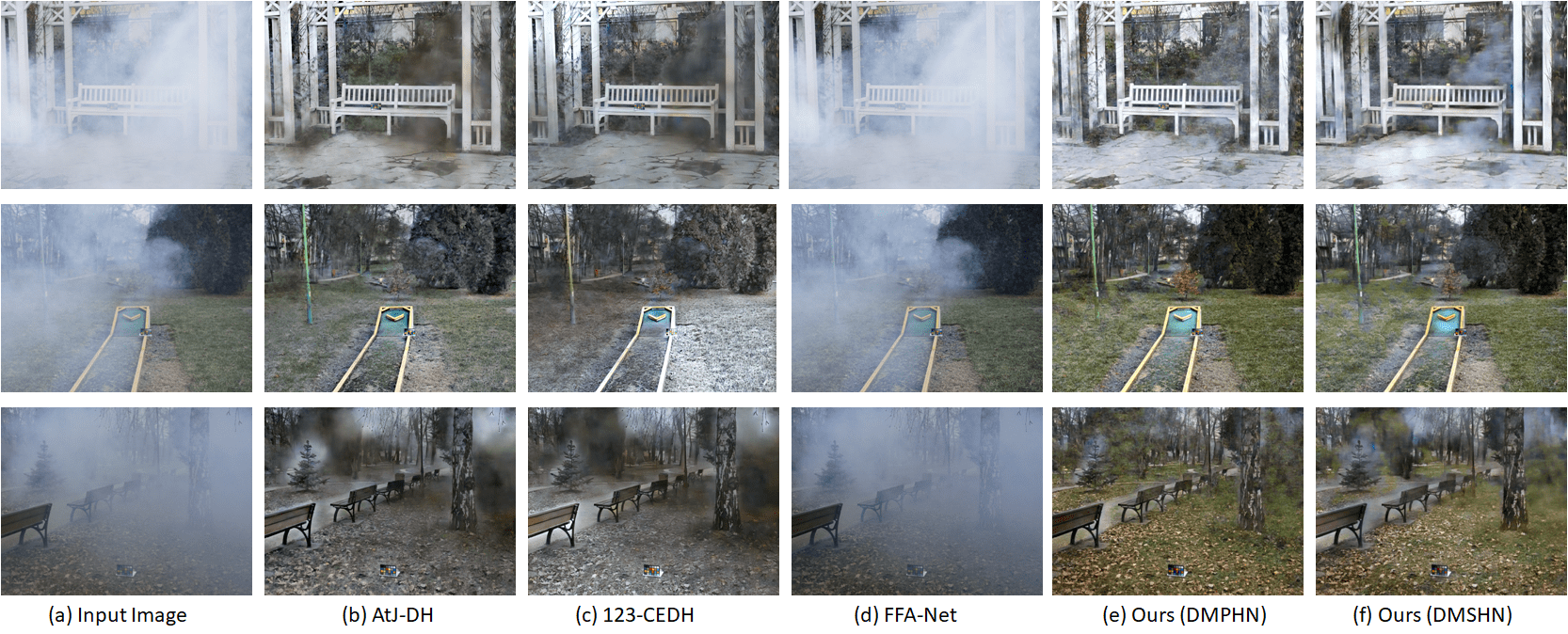}
    \caption{Qualitative results on NH-HAZE\cite{Ancuti_NH-HAZE_2020} Validation dataset.}
    \label{fig:my_label}
\end{figure*}
\subsubsection{NTIRE 2020 challenge on NonHomogeneous Image Dehazing:}
We participated in NTIRE 2020 challenge on NonHomogeneous Image Dehazing\cite{Ancuti_NTIRE_Dehaze_2020}. 27 teams submitted results in test phase, out of which 19 teams don't take help of extra training data like Dense-Haze\cite{densehaze,Ancuti_NTIRE_2019} and OHaze\cite{ohaze,Ancuti_NTIRE_2018}. The test results were evaluated on Fidelity measures as well as Perceptual Measures. Fidelity measures included PSNR and SSIM\cite{ssim-paper}, where LPIPS\cite{lpips}, Perceptual Index(PI)\cite{perc_index} and Mean Opinion Score(MOS) were used as Perceptual metrics. For fair comparison, we note down performances of some submissions that used only NH-HAZE dataset in Table-\ref{ntire_table}. 
% Our team \verb|Neuro_avengers|'s solution(DMPHN) is the fastest among all the submissions.
Our DMPHN network produced moderate quality outputs both in Fidelity and Perceptual metrics. Our network is the fastest entry among all the submissions.

% \usepackage{multirow}
\begin{table*}[h]
\centering
% \usepackage{multirow}
\begin{tabular}{|c|c|c|c|c|c|c|c|}
\hline
\multirow{2}{*}{Team} & \multicolumn{2}{c|}{Fidelity} & \multicolumn{3}{c|}{Perceptual quality} & \multirow{2}{*}{Runtime(s)$\downarrow$} & \multirow{2}{*}{GPU/CPU} \\ \cline{2-6}
                      & PSNR$\uparrow$           & SSIM$\uparrow$         & LPIPS$\downarrow$        & PI$\downarrow$           & MOS$\downarrow$       &                          &                          \\ \hline
method1               & 21.60          & 0.67         & 0.363        & 3.712        & 3         & 0.21                     & v100                     \\ \hline
method2               & 21.91          & 0.69         & 0.361        & 3.700        & 4         & 0.22                     & v100                     \\ \hline
method3               & 19.25          & 0.60         & 0.426        & 5.061        & 12        & 12.88                    & v100                     \\ \hline
method4               & 18.51          & 0.68         & 0.308        & 2.988        & 12        & 13.00                    & n/a                      \\ \hline
Ours (DMPHN)          & \textit{18.24}          & \textit{0.65}         & \textit{0.329 }       & \textit{3.051}        & \textit{14 }       & \textit{0.01}                     & 1080                   \\ \hline
method5               & 18.70          & 0.64         & 0.328        & 3.114        & 14        & 10.43                    & 1080ti                   \\ \hline
method6               & 18.67          & 0.64         & 0.303        & 3.211        & 16        & 1.64                     & TitanXP                  \\ \hline
method7               & 17.88          & 0.57         & 0.378        & 2.855        & 16        & 0.06                     & n/a                      \\ \hline
no processing         & 11.33          & 0.42         & 0.582        & 2.609        & 20        &                          &                          \\ \hline
\end{tabular}

\caption{NTIRE 2020 Nonhomogeneous challenge\cite{Ancuti_NTIRE_Dehaze_2020} Leaderboard. Submissions are sorted in ascending order of MOS.}
\label{ntire_table}
\end{table*}
\subsubsection{Dense Haze Removal:} 
\begin{figure*}
    \centering
    \includegraphics[width= \textwidth]{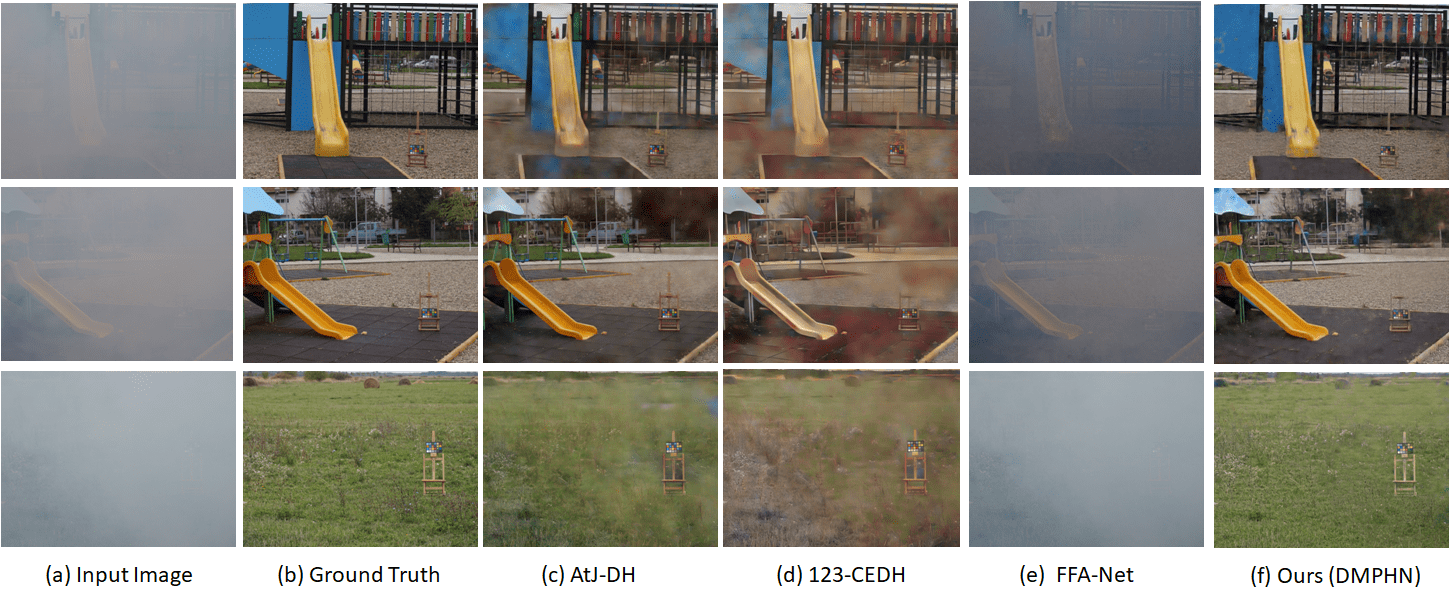}
    \caption{Qualitative results for Dense Haze Removal.}
    \label{dense_comp_pic}
\end{figure*}

DMPHN is effective for dense haze removal as well. We trained our network on Dense-HAZE dataset\cite{densehaze}. We train on 50 images for training and use 5 images for test. We compare the performance with  AtJ-DH\cite{atjdh}, 123-CEDH\cite{123cedh} and FFA-Net\cite{qin2019ffa}. Quantitative results and GPU runtimes are shown in Table-\ref{dense_comp}. We observe that DMPHN is significantly better than other models both in terms of fidelity measures and runtime. Figure-\ref{dense_comp_pic} shows qualitative comparison with the said models.
\begin{table}[!h!t!p]
\begin{tabular}{|c|c|c|c|}
\hline
            & PSNR  & SSIM   & Runtime(s) \\ \hline
AtJ-DH\cite{atjdh}      & 22.54 & 0.6436 & 0.0775     \\ \hline
123-CEDH\cite{123cedh}    & 19.63 & 0.5758 & 0.0559     \\ \hline
FFA-Net\cite{qin2019ffa}    & 11.93 & 0.3790 & 1.7472     \\ \hline
Ours(DMPHN) & \textbf{23.41} & \textbf{0.6669} & \textbf{0.0145}     \\ \hline
\end{tabular}
\centering
\caption{Quantitative Comparison on Dense-HAZE\cite{densehaze}.}
\label{dense_comp}
\end{table}

\subsection{Conclusion}
In this paper, we use a Multi-Patch and a Multi-Scale architecture for Nonhomogeneous haze removal from images. We show that DMPHN is better than DMSHN because DMPHN aggregates local features generated from a finer level to coarser level. 
Moreover, DMPHN is a fast algorithm and can dehaze images from a video sequence in real-time. We also show that DMPHN performs well for Dense Haze Removal. In future, the effectiveness of DMPHN with more levels can be explored for performance improvement, but the addition of more levels to architecture will subject to sacrifice in runtime.

\newpage

{\small
\bibliographystyle{ieee_fullname}
\bibliography{egpaper_for_review}
}

\end{document}